\documentclass[sigconf,nonacm, natbib=false]{acmart}
\usepackage{multirow}
\usepackage{graphicx}
\usepackage{caption}
\usepackage{float}
\usepackage{multicol}
\usepackage{tabularx}
\usepackage[singlelinecheck=off]{caption
}
\usepackage{url}
\usepackage[utf8]{inputenc}

\AtBeginDocument{%
  \providecommand\BibTeX{{%
    \normalfont B\kern-0.5em{\scshape i\kern-0.25em b}\kern-0.8em\TeX}}}

\copyrightyear{2023}
\acmYear{2023}
\setcopyright{rightsretained}
\acmConference[MLLD '23]{October 22, 2023}{Birmingham}{United Kingdom} %

\acmPrice{15.00}
\acmISBN{978-1-4503-XXXX-X/18/06}

\RequirePackage[
  datamodel=acmdatamodel,
  style=acmnumeric,
  ]{biblatex}

\begin{document}

\title{LEEC: A Legal Element Extraction Dataset with an Extensive Domain-Specific Label System}




\author{Zongyue Xue}
\authornote{Equal contribution. Listing order is random.}
\email{xuezy21@mails.tsinghua.edu.cn}
\affiliation{%
  \institution{Law School, Tsinghua University}
  \country{Beijing, China}}

\author{Huanghai Liu}
\authornotemark[1]
\email{liuhh23@mails.tsinghua.edu.cn}
\affiliation{%
  \institution{Law School, Tsinghua University}
  \country{Beijing, China}}

\author{Yiran Hu}
\authornotemark[1]
\authornote{Corresponding}
\email{huyr21@mails.tsinghua.edu.cn}
\affiliation{%
  \institution{Law School, Tsinghua University}
  \country{Beijing, China}}

\author{Kangle Kong}
\email{kkl22@mails.tsinghua.edu.cn}
\affiliation{%
  \institution{Law School, Tsinghua University}
  \country{Beijing, China}}

\author{Chenlu Wang}
\email{wangchenlu2014@gmail.com}
\affiliation{%
  \institution{Law School, Tsinghua University}
  \country{Beijing, China}}
  \email{}

\author{Yun Liu}
\authornotemark[2]
\email{liuyun89@tsinghua.edu.cn}
\affiliation{%
  \institution{Law School, Tsinghua University}
  \country{Beijing, China}}

\author{Weixing Shen}
\authornotemark[2]
\email{wxshen@tsinghua.edu.cn}
\affiliation{%
  \institution{Law School, Tsinghua University}
  \country{Beijing, China}}



%
\renewcommand{\shortauthors}{Xue and Liu, et al.}

\begin{abstract}
As a pivotal task in natural language processing, element extraction has gained significance in the legal domain. Extracting legal elements from judicial documents helps enhance interpretative and analytical capacities of legal cases, and thereby facilitating a wide array of downstream applications in various domains of law. Yet existing element extraction datasets are limited by their restricted access to legal knowledge and insufficient coverage of labels. To address this shortfall, we introduce a more comprehensive, large-scale criminal element extraction dataset, comprising 15,831 judicial documents and 159 labels. This dataset was constructed through two main steps: first, designing the label system by our team of legal experts based on prior legal research which identified critical factors driving and processes generating sentencing outcomes in criminal cases; second, employing the legal knowledge to annotate judicial documents according to the label system and annotation guideline. The Legal Element ExtraCtion dataset (LEEC) represents the most extensive and domain-specific legal element extraction dataset for the Chinese legal system. Leveraging the annotated data, we employed various SOTA models that validates the applicability of LEEC for Document Event Extraction (DEE) task. The LEEC dataset is available on \href{https://github.com/THUlawtech/LEEC}{https://github.com/THUlawtech/LEEC}.
\end{abstract}


\maketitle

\section{Introduction}

Extracting key elements of judicial documents and their relations is valuable for analyzing legal cases and making sentencing decisions. Meanwhile, the disparity between “law in books” and “law in action” introduces a significant difficulty in fully capturing the important elements in judicial practice, thereby augmenting the complexity of element extraction. With the help of the extensive label system constructed on the legal knowledge graph by our team of legal experts, Legal Element ExtraCtion (LEEC) dataset aims to provide element mentions, trigger words and values munually annotated from large-scale judicial documents with high quality. This could facilitate automatic extraction of elements, benefiting numerous LegalAI applications, such as Legal Judgement Prediction and Similar Case Retrieval, as well as empirical legal research. Meanwhile, with an extensive label system based on prior empirical legal research, LEEC could provide comprehensive labels that are important in judicial practice yet neglected by prior studies regarding element extraction, while also contribute to the replication and innovation in empirical studies. 

Inspired by the success of general-domain element extraction\cite{hogenboom2011overview} \cite{liao2010using}\cite{guo2020knowledge}, previous studies\cite{shen-etal-2020-hierarchical}\cite{feng-etal-2022-legal}\cite{sierra2018event} attempted to construct an element extraction system in the legal domain, leveraging both hand-crafted features and neural networks. For instance, LeCaRD\cite{DBLP:conf/sigir/MaSW000M21}, the first Legal Case Retrieval Dataset in China contains 107 query cases and 10,700 candidate cases selected from of over 43,000 Chinese criminal judgements, was constructed. LEVEN is a large-scale Chinese Legal event detection dataset\cite{yao2022leven}, with 8,116 legal documents and 150,977 human annotated event mentions in 108 event types. At present, the existing datasets in China also includes CAIL\cite{DBLP:journals/corr/abs-1911-08962}, Criminal\cite{DBLP:conf/sigir/MaSW000M21}, CJO\footnote{\url{https://wenshu.court.gov.cn}}, PKU\footnote{\url{https://home.pkulaw.com/}}, etc. However, there are several main challenges in the existing work:


(1) \textbf{Incomprehensive Label System.} Existing label systems\cite{lithuircoliee} \cite{li2023sailer}\cite{li2023thuircoliee}\cite{richards2016explaining} of prior studies mainly lay emphasis on a limited scope of charge-oriented elements. The current element schema, especially those in the Chinese contexts, are far from enough. For example, victim number in highly likely a salient predictor in crime type and sentencing in Chinese criminal trials, yet we did not find any prior study in Chinese contexts that incorporate this label in their label system. Besides, existing studies predominantly focus on the legally prescribed factors in sentencing, overlooking extra-legal elements. However, a wealth of empirical research suggests that these elements, such as the defendant's and victim's age, gender, race/ethnicity, etc., may significantly influence trial and sentencing outcomes \cite{richards2016explaining}\cite{doerner2010independent}\cite{ulmer2012recent}\cite{chen2023equals}\cite{tran2019building}. The absence of these factors in the label system may compromise the performance of downstream tasks.

(2) \textbf{Lack of Domain Focus.} An overwhelming majority of existing datasets\cite{li2020duee}\cite{nguyen2016dataset}\cite{veyseh2022mee} for element extraction mainly focus on the element or event extraction in the general domain. However, such datasets may not be well suited to applications in the legal domain. For example, Recidivist (\textit{Leifan} in Chinese) and Previous Criminal Record (\textit{Qianke} in Chinese) are closely-connected yet distinct legal concepts in Chinese criminal law, which could be difficult to distinguish without adequate legal knowledge. Furthermore, various court participants may present different interpretations and perspectives on the same legal elements, such as whether the defendant voluntarily surrendered, confessed, or pled guilty. This diversity of opinion can cause confusion without professional legal annotation. Therefore, existing datasets from general domains are hardly applicable for comprehensive analysis and tasks based on legal texts owing to their lack of adequate understanding of legal knowledge and contexts.

(3) \textbf{Inadequate Few-Shot Charge Coverage.} Existing datasets\cite{liu2023investigating} predominantly focus on high-frequency charges and often underperform when dealing with less frequent charges due to the limited number of cases. Moreover, charges that typically manifest with similar descriptions, such as the crime of forcible seizure and the crime of robbery, can be challenging to differentiate, especially with limited data.

(4) \textbf{Limited Application.} The previous datasets\cite{yao2023unsupervised} in the legal domain do not consider the relations of elements, hindering the comprehension of judicial documents as they frequently present complex relationships. For example, in a document involving multiple defendants, the extracted labels of one defendant may not be applicable to another defendant, and thus, these labels should be linked to their corresponding defendant. The oversight of relational context frequently appear in judicial practice may significantly limit the performance of downstream tasks based on these datasets and their practical utility, including their usage in empirical analysis and effective LegalAI applications in real-world court settings.

To provide a solid foundation for legal element extraction, LEEC alleviates the above limitation in the following way: 

(1) \textbf{Extensive Label System.} Our team of legal experts not only extended the coverage of legally prescribed factors that may significantly impact Chinese criminal trials and sentencing, but also actively drew upon a comprehensive collection of Chinese empirical studies published in Chinese core journals, as well as important legal papers in Chinese contexts published internationally. Based on the theory, research design, and findings of these empirical studies, we systematically compiled extra-legal key labels regarded by these studies as having substantial impact in Chinese judicial practice. In this way, we are able to constructed an extended and comprehensive label system in the legal domain. 

(2) \textbf{Large Scale.} LEEC is annotated based on the publicly available cases of both LEVEN and LeCaRD, with a total of 15,831 cases. Therefore, the high coverage of cases could largely alleviate the problem of limited number of cases in few-shot charges, leading to an increased ability of meeting needs in real-world court settings. Besides, the annotation of LEEC could be combined with the previous annotation from LEVEN and LeCaRD, providing more comprehensive information to facilitate the analysis of judicial documents. 

(3) \textbf{Broader Application.} The knowledge graph we developed for our annotation system encapsulates significant relationships among various elements. For instance, as it is frequent for a single Chinese judicial document to involve multiple defendants, crimes, and victims, our team of legal experts has effectively linked defendant and victim characteristics to their respective individuals and affiliated crime characteristics to the corresponding offenses. This integration of crucial interrelations among labels enhances performance in various downstream applications, including the prediction of a specific defendant's crime and sentencing, and also expands LEEC's applicability in real-world court settings and future empirical research.

To validate the quality and applicability of LEEC, we implement various SOTA models in the field of document-level event extraction and evaluate them on our dataset, which shows that the elements of of LEEC could be extracted with relatively high accuracy by these models.

\section{Data analysis}

\subsection{Corpus and Preprocessing}
Our selection of cases draws from the publicly available LEVEN\cite{yao2022leven}  and LeCaRD\cite{DBLP:conf/sigir/MaSW000M21}datasets. These cases undergo a preprocessing phase wherein we extract the full text of each case, as well as their respective case number, crime, and year of judgement using automated algorithm, to facilitate the subsequent manual annotation. This complete dataset is comprised of 17,352 cases. After deduplication, the number of total unique cases is 17,231, encompassing 10,805 cases from LeCaRD, and 6,426 cases from LEVEN. All documents within this collection represent criminal decisions delivered in the span of the past two decades. Among the complete dataset, we preserve 1400 judicial document as non-published test dataset for future evaluation, and publish the remaining annotated data from 15,831 judicial documents. We compare LEEC with two types of datasets: (1) General-domain ED datasets. Compared with ACE2005\cite{grishman2005nyu} and MAVEN \cite{wang2020maven}, LEEC is an element extraction dataset in the field of Chinese criminal law, with a label system designed for legal texts. The judicial documents of LEEC are also annotated by law school students with adequate understanding of legal knowledge and concepts. (2) Legal-domain datasets. Compared with LEVEN and LeCaRD, our dataset offers a comprehensive expansion of the label system with a finer level of granularity, encapsulating both legal and extra-legal labels. Furthermore, LEEC encompasses independent annotations of distinct entities such as victims, defendants, and crimes. This methodological approach considerably enhances the dataset's precision and applicability, thereby contributing significantly to downstream applications and empirical research.

\subsection{Data Distribution}

Unlike previous datasets, LEEC incorporates information on multiple victims, defendants, and causes. It should be noted that a minor fraction of the annotated victim number contains missing values. This is attributable to instances where the precise number of victims cannot be ascertained due to the insufficient information provided within the judicial document. For a detailed explanation concerning the occurrence of missing values within the annotation, please see the Annotation section. Our dataset reveals the presence of multiple defendants in 34\% of cases, multiple victims in 19\% of all cases, and 40\% of cases that contain at least one victim, and multiple crimes in 43\% of cases. This underscores the necessity and effectiveness of introducing a sophisticated, domain-specific label system to handle such complexity. The distribution of cases is displayed in Table 1.
\begin{table}[H]
	\renewcommand{\arraystretch}{1.5} 
	\resizebox{0.8\linewidth}{!}{ 
	\begin{tabular}{l|l|l|l|l|l|l}
		\hline
		\hline
		\textbf{Defendant number} & \textbf{1} & \textbf{2}	& \textbf{3} & \textbf{4} &\textbf{5} &\textbf{>5}\\
		\hline
		\textbf{Case distribution}& \textbf {10465} & \textbf {2262} &\textbf {1182} & \textbf {644} & \textbf {406}& \textbf {872}\\
		\hline
		\hline
  		\textbf{Victim number} & \textbf{0} & \textbf{1}	& \textbf{2} & \textbf{3} &\textbf{4} &\textbf{>4}\\
		\hline
		\textbf{Case distribution}& \textbf {8140} & \textbf {4614} &\textbf {954} & \textbf {567} & \textbf {329}& \textbf {1226}\\
		\hline
        \hline
        \textbf{Crime number} & \textbf{1} & \textbf{2}	& \textbf{3} & \textbf{4} &\textbf{5} &\textbf{>5}\\
		\hline
		\textbf{Case distribution}& \textbf {9066} & \textbf {2920} &\textbf {1312} & \textbf {747} & \textbf {463}& \textbf {1323}\\
		\hline
		\hline
	\end{tabular}}
\caption{\centering \textbf{Case distribution on multi-defendants, victims, and crimes.}}
	\label{table:3}
\end{table}

\section{Label System}
This section encompasses the compilation of the extensive label system, as well as the establishment of the crucial relationships among those labels.
\subsection{Label Compilation}
Our team of legal experts, led by professors in law, incorporated a wide range of legal and extra-legal elements to build a comprehensive and extended knowledge graph covering key elements within the Chinese legal domain. First, our team of legal experts compiled the crucial legal circumstances and factors stipulated by Chinese criminal law and legal interpretations, such as whether the defendant confessed, pled guilty, voluntarily surrendered, conducted justifiable defense, etc.. Furthermore, it has long been underscored by researchers that there is to cope with the distinction between “law in books” and “law in action”, underscoring that the complexity of the actual application of law and its potential divergence from the written text\cite{pound1910law}. It has widely been revealed that extra-legal factors may significantly impact the application of law, including judicial decisions, in practice. Therefore, we utilized elements and theories developed and validated by empirical legal research to comprehensively capture the important factors in Chinese criminal trials. 

Specifically, our team of legal experts systematically compiled 178 quantitative legal studies from 2018 to 2022 published across 22 journals in Chinese listed in the China Legal Science Citation Index (CLSCI), an index curated by the Law Institute of China Law Society, providing the list of core legal journals in China. The majority of these studies investigated the impact of various legal and extra-legal factors on sentencing based on publicized judicial documents. As such, the labels and theories used in these studies serve as a valuable source of factors that may significantly influence judicial decisions, thereby facilitating downstream tasks. In addition, we drew upon a wide range of empirical legal studies published in SSCI journals, especially those investigating sentencing factors in Chinese contexts. Our team meticulously collected the core theories and labels used in these studies and incorporated them into our legal system. 

For instance, the Group Threat Theory suggests that when majority groups feel threatened by minority populations, criminal justice systems may treat racial or ethnic minorities adversely\cite{ulmer2004sentencing}. This theory has also been validated and developed in the Chinese context by prior empirical research, which found that minorities perceived as “problem minorities” that might disrupt public order may face discrimination in Chinese criminal cases\cite{hou2020ethnic}. Therefore, we included the ethnic status of offenders in our knowledge graph. Moreover, the Focal Concern Theory highlights the strategic function of judges in contexts of managerial uncertainty and constrained knowledge. It identifies three crucial factors influencing sentencing decisions: the defendant’s culpability, the risk posed to the community, and pragmatic considerations such as the court’s workload\cite{steffensmeier2000ethnicity}. Research in Chinese contexts has shown that, in line with the Focal Concern Theory, defendant’s being a rural-to-urban migrant – measured by his or her registered permanent residence (\textit{Hukou}) – significantly impacts sentencing outcomes\cite{jiang2018hukou}. As a result, we also included the registered permanent residence of defendants as an important element in our knowledge graph. Meanwhile, it should be noted that some elements could be easily and accurately extracted through identifying keywords or regular expression matching in Chinese criminal verdicts, and thus, do not require manual annotation, such as court name, judge name, case title, year of judgement, etc.. Therefore, these elements are not annotated or included in our knowledge graph and label system. Following this scheme, we effectively constructed an extended, multi-level knowledge graph to cover 159 important elements – both legal and extra-legal – in Chinese criminal sentencing. The elements in the knowledge graph are divided into four main categories: defendant characteristics, victim characteristics, case characteristics, and crime characteristics.

\subsection{Relation Construction}
We integrated the relations among elements into the knowledge graph we constructed, recognizing their significant influence on judicial decisions. This integration is particularly important in verdicts involving multiple crimes, offenders, victims, or defenders. To illustrate, consider a verdict with several defendants: the circumstances and characteristics of one defendant might differ from those of the others. Therefore, all characteristics pertaining to a victim or defendant are linked directly to the relevant individual. Besides, considering a defendant may have up to two defenders in Chinese criminal trials, the defender characteristics are also connected to the individual defender of a specific defendant. Furthermore, all characteristics of a crime are associated with the specific crime committed by a particular defendant. This is imperative because each defendant could be sentenced for multiple crimes in Chinese judicial documents. However, the circumstances of a specific crime may not necessarily apply to another. It is also noteworthy that in Chinese criminal cases where an individual defendant committed multiple crimes, the court typically adjudicates a sentence for each individual crime, followed by an overall aggregated sentence. This final sentence, which is usually subject to a certain degree of the judge's discretion, may not necessarily align with the sum of the individual sentences. Consequently, in our knowledge graph, we deliberately included both the sentencing elements, linked to each distinct crime of a specific defendant, and the final, aggregated sentence, linked to each defendant. The elements within the knowledge graph is depicted in Figure 1.

\section{Annotation}
The annotation of LEEC requires the annotators to find and determine the element mentions, trigger words, values of each elements from the documents. Specifically, the annotation is conducted manually by a team of graduates and undergraduates majoring in law, trained and led by professors in law. All of them are interviewed before joining the team to ensure their ability to comprehend Chinese legal concepts and knowledge, and practiced for several hours before formal annotating. We have compiled a comprehensive 155-page annotation guideline in Chinese to assist our annotators. It provides an in-depth understanding of each element and their respective annotation methods. The guideline includes definitions, potential values for each element, common locations within the judicial documents where the element frequently appears, detailed rules for annotation, and real-world examples of document annotation for the elements, etc.. The examples of the annotation guideline are provided in detail in Appendix A.

During annotation, we adopted a two-stage process. In the first stage, we performed a fine annotation of 4182 documents randomly selected from the public datasets of LeCaRD and LEVEN. We annotated the element mentions, trigger words, and the values of each element, which are the basis for evaluating several baselines of the document-level event extraction task to validate the quality and applicability of the annotations for extraction tasks. The second stage is the extended annotation of the values of elements on the remaining public datasets of LeCaRD and LEVEN, which covers 11624 documents, to assist the verification of the results of element extraction tasks. Besides, as the label system of LEEC is largely originated from prior empirical studies, this extended dataset could also contribute to the replication and exploration of empirical research. In both stages, we performed double annotations on a portion of the work of each annotator to check the consistency and quality of the annotation. Datasets from both stages are publicly available on \href{https://github.com/THUlawtech/LEEC}{https://github.com/THUlawtech/LEEC}. 

During the annotation process, we observed that a minor fraction of judicial documents contained an unusually high number of defendants or victims, in some cases reaching into the hundreds. These documents were predominantly associated with corporate crimes committed for financial gains. In response to this situation, we implemented a upper limit, treating cases with more than seven defendants or victims as if they had exactly seven, and only annotated the first seven defendants and victims.

It is also noteworthy that a number of elements may not always be explicitly mentioned in judicial documents. For instance, the offender gender, while frequently disclosed, is not always explicitly stated, as discretion is commonly exercised in such circumstances. Similarly, whether an offender has received forgiveness from victims or their close relatives is sometimes clearly affirmed or negated, yet such information may be not mentioned in many judicial documents. As a general rule, we classify these non-mention instances as missing values. However, an exception exists when our team of legal experts determines that, in almost all cases in Chinese judicial practice, a certain element (of binary nature) is specified in the judicial document when it holds a value of 1. Thus, its absence from a judicial document suggests that the element holds a value of 0. An example is the COUNTERCLAIM element, which denotes whether the defendant has lodged a counterclaim against the private prosecutor or the victim in the incidental civil action portion of the criminal case. In practice, this element is typically documented if and only if a counterclaim has indeed been initiated. Hence, should the element be absent from the judicial document, the annotator would assign it a value of 0. The same principle applies to elements concerning the sentencing type of offenders. If the offender does not receive a specific type of punishment, the judicial document would not mention the corresponding element in the sentencing section. Therefore, such non-mention instances are also assigned a value of 0 during annotation. Tables 5 through 15 contain special notes for elements wherein non-mention does not equate to missing value. 

We measure the data quality by Kappa, with a value of 0.71. This value demonstrates that the manual annotation of LEEC is conducted with high quality, contributing to the development of legal element extraction and the analysis of legal cases.

\section{Document-level Event Extraction}
This section delves into the specifics of the experiment conducted using the LEEC dataset. It covers the experiment's background and settings, the suitable baselines and metrics, as well as the results and associated discussions.

\subsection{Experiment Settings}
For the DEE task, we selected some representative labels in LEEC labeling system to extract important event information of the defendant, and the parameters in the event table are shown in Table 2. Since most legal cases do not typically demonstrate a direct correspondence relation between defendants and victims when there are multiple defendants and victims involved, we uniformly assign the name of the first victim appearing in the document as the victim's name. The maximum number of defendants to be drawn from a judgment document is 7. After the above filtering steps, the LEEC-DEE dataset size is 4156, with 524 annotated documents from LEVEN and 3632 annotated documents from LeCaRD. We split the dataset into the training, validation and test sets at a proximate ratio of 8:1:1.
We use the same vocabulary as \cite{zheng2019doc2edag}and randomly initialize all the embeddings where dh=768 and dl=32. We employ the Adam optimizer with the learning rate 2e-5 and the batchsize is 64. All models are trained for 100 epochs and the checkpoints with the best F1 scores on the dev set are selected for evaluation on the test set.

\begin{table}[H]
	\renewcommand{\arraystretch}{1.2} 
	\resizebox{\linewidth}{!}{ 
	\begin{tabular}{l|l|l}
		\hline \hline
		\textbf{Defendant}                                & \textbf{Role Type}  & \textbf{Corresponding label}                                                      \\ \hline
		\multirow{5}{*}{\textbf{Demographic Characteristics}}     & Name                & Defendant\_name                                                               \\ \cline{2-3} 
		& Gender & Defendant\_gender                                                  \\ \cline{2-3} 
		& Birth & Defendant\_birth                                                  \\ \cline{2-3} 
		& Nation              & Defendant\_nationality                                    \\ \cline{2-3} 
		& Place               & Defendant\_birthplace                                 \\ \hline		
  \multirow{16}{*}{\textbf{Aggregated Sentencing}} & Control  & Probation\_aggregated                                    \\ \cline{2-3} 
		& ControlTime         & \multicolumn{1}{l}{Probation\_term\_aggregated}                                          \\ \cline{2-3} 
		& Detention           & \multicolumn{1}{l}{Limited\_incarceration\_aggregated}                       \\ \cline{2-3} 
		& DetentionTime       & \multicolumn{1}{l}{Limited\_incarceration\_term\_aggregated}                               \\ \cline{2-3} 
		& Imprisonment        & \multicolumn{1}{l}
        {Fixed\_imprison\_aggregated}         \\ \cline{2-3} 
		& ImprisonmentTime    & \multicolumn{1}{l}{Fixed\_imprison\_term\_aggregated}                          \\ \cline{2-3} 
		& PoliticalRights     & \multicolumn{1}{l}{Political\_deprivation\_aggregated} \\ \cline{2-3} 
		& PoliticalRightsTime & \multicolumn{1}{l}{Political\_deprivation\_term\_aggregated}                  \\ \cline{2-3} 
		& Fine                & \multicolumn{1}{l}
        {Fine\_aggregated}                                   \\ \cline{2-3} 
		& FineNum             & \multicolumn{1}{l}{Fine\_amount\_aggregated}                                             \\ \cline{2-3} 
		& PartofProperty      & \multicolumn{1}{l}{Partial\_property\_confiscation\_aggregated}                 \\ \cline{2-3} 
		& PartofPropertyNum   & \multicolumn{1}{l}{Partial\_confiscated\_amount\_aggregated}                             \\ \cline{2-3} 
		& AllProperty         & Total\_property\_confiscation\_aggregated                                             \\ \cline{2-3} 
		& AllPropertyNum      & Total\_confiscated\_amount\_aggregated                                          \\ \cline{2-3} 
		& EcoCompensation     & Loss\_compensation\_aggregated                \\ \cline{2-3} 
		& EcoCompensationNum  & Compensation\_amount\_aggregated \\ \hline
		\textbf{Victim information}                       & VictimName          & Victim\_name                                                              \\ \hline \hline
	\end{tabular}}
\caption{\centering\textbf{Roles in the defendant event table and their corresponding labels in the LEEC label system.}
}
\label{table:1}
\end{table}

\subsection{Baselines and Metrics}
Baselines. We introduce the following typical models as baselines: (1) DCFEE\cite{yang2018dcfee} is the first model introduced Distance Supervision (DS) to solve DEE task. there are two variants included: DCFEE-O only extracts one event record from one document while DCFEE-M tries to extract multiple possible event records. (2) Doc2EDAG\cite{zheng2019doc2edag} is an end-to-end DEE model that constructs event records in an auto-regressive way by generating entity-based Directed Acyclic Graphs (DAGs). (3) GreedyDec is a baseline proposed in Doc2EDAG\cite{zheng2019doc2edag} which fills one event table greedily. (4) PTPCG\cite{zhu2021efficient} is a lightweight model for end-to-end document-level event extraction based on pruned complete graphs with pseudo triggers.

Metrics. Following the same evaluation setting in the previous studies\cite{zheng2019doc2edag}\cite{zhu2021efficient}. For each prediction record, we select a golden record by matching records with the same event type and the most shared arguments, and calculate the F1 score by comparing the parameters between them.

\subsection{Results}
Table 3 shows the experimental results on LEEC-DEE dataset. Table 4 presents the the epochs where the baselines perform optimally. From experiments we have the following observations: 1) The ability of most models is consistent with their performance on previous DEE data-sets\cite{zhu2021efficient}; 2) Some baselines cannot converge well on datasets, such as \url{Doc2EDAG}\cite{zheng2019doc2edag} and similar structured GreedyDec. One reason is that compared to the previously evaluated data in the general or financial fields, the text of legal documents is longer and the arguments are more dispersed, which is not conducive to the Doc2EDA-G\cite{zheng2019doc2edag} structure using a sequential path extension method for reasoning. Experiments show that LEEC-DEE dataset poses challenges to DEE models, indicating that DEE in legal domain is an open issue.

\begin{table}[H]
	\renewcommand{\arraystretch}{1.5} 
	\resizebox{0.5\linewidth}{!}{ 
	\begin{tabular}{l|l|l|l}
		\hline
		\hline
		\textbf{Model} & \textbf{Precision} & \textbf{Recall}	& \textbf{F1 score} \\
		\hline
		\textbf{DCFEE-O}& 67.37\% & 77.90\% &72.25\% \\	
		\textbf{DCFEE-M}& 67.03\% &72.85\% &69.82\% \\			
		\textbf{Greedy-Dec} & 79.63\% & 59.25\%  &67.94\%\\
		\textbf{Doc2EDAG} & 26.61\% & 72.39\% &38.91\% \\
		\hline
		\hline
	\end{tabular}}
\caption{\centering\textbf{Overall performance on Document-level Event Extraction.}}
	\label{table:2}
\end{table}

\begin{table}[H]
	\renewcommand{\arraystretch}{1.5} 
	\resizebox{0.5\linewidth}{!}{ 
				\begin{tabular}{l|l|l}
					\hline
					\hline
					\textbf{Model} & \textbf{Train Epoch}	& \textbf{Best Epoch} \\
					\hline
					\textbf{DCFEE-S} & 100 & 24 \\	
					\textbf{DCFEE-M} & 100 & 15 \\			
					\textbf{Greedy-Dec} & 100 & 6 \\
					\textbf{Doc2EDAG} & 100 & 8 \\
					\textbf{PTPCG} & 100 & 88 \\
					\hline
					\hline
				\end{tabular}}
		\caption{\centering\textbf{The train epoch and best epoch in which the models achieve the highest micro F1 score on the dev set.}}
		\label{table:3}
\end{table}

\section{Discussion and Conclusion}

In this study, we introduce LEEC, a uniquely tailored dataset designed for the extraction of legal elements within the Chinese criminal law system. Our dataset stands out from existing element extraction datasets as it is enriched with an expansive legal domain label system, meticulously curated by our team of legal experts. This system integrates crucial legal knowledge drawn from Chinese law, empirical legal studies, and an understanding of the legal contexts derived from judicial practices. Each of the 15,831 cases in the dataset has been annotated by law school students. Experimental results underline the challenges associated with multi-label prediction, signifying an area of focus for future research.

It is pertinent to note that the knowledge graph and label system developed in this study exhibit a notably low level of granularity. This specificity was designed to support valuable downstream applications and empirical legal research. Yet, it is of paramount importance that users of LEEC exercise due caution. We vehemently oppose to the use of LEEC for any purposes that could lead to discrimination or violate the principle of the rule of law, whether within or outside the courtroom. The personal information included in the published judicial documents was collected and processed in strict compliance with Chinese law. Any future utilization of LEEC must also adhere to applicable laws and commit to responsible, ethical handling of this data to prevent misuse and uphold the principle of privacy. As it is crucial that all users understand their role in maintaining ethical standards and protecting personal information, we urge each user to consider the potential implications of their work and to use the dataset of this study responsibly.

This study has several limitations that we hope will be addressed by future research: 1) We selected only 22 elements for the DEE task as the corresponding text of these elements could be clearly and directly extracted from the document, such as the defendant's name. However, some elements within the system cannot always be directly extracted, such as the JOINT\_CRIME element, which can be expressed in a flexible manner within judicial documents. Consequently, the annotation of this element necessitates the specification of complex rules and, occasionally, a certain degree of reading comprehension. Developing models to extract such labels may present a significant challenge; 2) The study inherently deals with sparse matrix for some elements that appear infrequently in the judicial documents, such as the LEGAL\_AID element, which may pose challenges in the DEE task. The sparsity of data can lead to computational difficulties and poor model performance. Future research should consider strategies for handling and interpreting sparse data. 3) Based on estimations, approximately 75\% of all judicial verdicts in China were ultimately disclosed for cases not processed through criminal mediation in recent years \cite{tang2019mass}. This suggests that our dataset may inevitably contain selection bias. Therefore, the distribution and correlations of labels in our dataset may not fully represent those in actual Chinese courts. Future studies should consider this potential bias when interpreting findings based on this dataset.

\section*{Acknowledgements}

We extend our sincere gratitude to Huaiyu Hu and Senyu Li from the Law School at Tsinghua University for their invaluable legal assistance. Meanwhile, we wish to express our deep appreciation to Zexia Yang and Ziheng Xie, also from the Law School at Tsinghua University, for their help and advice regarding data processing and knowledge graph construction. This work is supported by the National Key Research and Development Program of China (No. 2022YFC3301504).


\clearpage
\printbibliography
\clearpage

\appendix
\section{Examples from the Annotation Guideline} 
\subsection{Joint\_crime}
\textbf{1.	Label Meaning\\}
A JOINT\_CRIME element refers to whether the court determined that an intentional crime was committed by two or more persons or units in collaboration.\\
\textbf{2.	Potential Value of Element\\}
1 = Yes; 0 = No\\
\textbf{3.	Annotation Rules\\}
(1) This element generally appears in the fact description section (typically after the statement of “The court finds”) or the reasoning section (typically after the statement of “The court considers”).\\
(2) This element cannot be determined solely based on the number of defendants that appear in the judicial document. Having multiple defendants does not necessarily constitute a joint crime; having only one defendant does not necessarily mean there is no joint crime.\\
(3) If the judicial document explicitly states that “the behavior is a joint crime”, then it is a joint crime. If there are other defendants mentioned in the same criminal act, but handled separately, it also counts as a joint crime.\\
(4) If terms like “in league with others” or “conspiring with others” appear, it is generally considered a joint crime. Any ambiguities or uncertainties should be reported to us.\\
(5) If the document does not explicitly state that the crime is a joint crime, but contains phrases such as the defendant “is an accomplice”, “is a principal offender”, it is considered a joint crime.\\
(6) If the joint crime status is inconsistent across different charges, each crime should be annotated separately.\\
(7) If it can be clearly determined from the judicial document that there is only one perpetrator, it can be determined that it is not a joint crime.\\
(8) If a crime contains multiple criminal acts, but only some of the criminal facts are jointly committed, the rest are not, it may still be considered a joint crime.\\
(9) In corporate crimes, the entity and its directly responsible personnel may be both considered guilty by the court, but this does not constitute a joint crime.\\
\textbf{4.	Annotation Example\\}
\textbf{Original Document Text\\}
Case number: (2011) Yong Zhen Xing Chu Zi No.49\\
From November to December 1994, the defendant FANG Xingdu and FANG Jinqi (already sentenced) \textbf{conspired in advance}, using the convenience of FANG Xingdu's position as a guard at the original Ningbo Heqiao Chemical Co., Ltd. (now Ningbo Xinqiao Chemical Co., Ltd.), responsible for the receipt of styrene raw materials, when FANG Jinqi drove Ningbo Chemical Hazardous Goods Transport Company's chemical tank truck to transport styrene raw materials from Zhenhai Port Area to Ningbo Heqiao Chemical Co., Ltd... The court believes that the defendant FANG Xingdu \textbf{conspired with others}...\\
\textbf{Annotation Result\\}
1\\
\textbf{Reason for Annotation\\}
“Conspired in advance” indicates that the two defendants conspired with each other in advance; “conspired with others” is also a typical expression of joint crime. Therefore, even if the judgment document does not explicitly state that the defendant committed a “joint crime”, it can also be marked 1.

\subsection{Forgiveness}
\textbf{1. Label Meaning}\\
A FORGIVENESS element refers to whether the court determined that the defendant had obtained forgiveness from victims or their close relatives.\\
\textbf{2. Potential Value of Element\\}
1 = Yes; 0 = No\\
\textbf{3. Annotation Rules\\}
(1) This element generally appears in the fact description section (typically after the statement of “The court finds”) or the reasoning section (typically after the statement of “The court considers”).\\
(2) For this element to be assigned a value of 1, the document usually contains expressions such as “obtained the forgiveness of the victim's relatives” or “obtained the forgiveness of the victim”. The forgiveness here includes the victims and their close relatives. \\
(3) The element should be annotated for each defendant who appears in the judicial document, respectively.\\
(4) If there are multiple defendants in the case, and one of them obtains forgiveness, it does not mean that the victim also forgives other defendants.\\
(5) If there are multiple victims in the case, and only some of the victims forgave the defendant, this still constitutes the circumstance of obtaining forgiveness, and thus, the element should be assigned a value of 1.\\
\textbf{4. Annotation Example}\\
\textbf{Original Document Text\\}
Case Number: (2017) Yu 1381 Criminal First Instance 426\\
The defendants, WANG Congwen and HUANG Jinlian, confessed their crimes truthfully, and thus, could receive lighter punishments. They compensated the economic losses of the victim's close relatives, and \textbf{obtained the forgiveness of the victim's close relatives}. Therefore, they can be punished lightly at discretion. The defendant HUANG Jinlian has shown remorse and has no danger of reoffending. A suspended sentence has no significant adverse impact on the community where she lives...\\
\textbf{Annotation Result}\\
1\\
\textbf{Reason for Annotation}\\
The document clearly states that the defendants obtained the forgiveness of the victim's relatives.\\

\section{Element Schema and Description} 
To facilitate the understanding of our label system and each element within it, thereby promoting future application and research, we provide the multi-level label system in Figure 1. Additionally, Tables 5 to 15 provide detailed descriptions of each element, including element name, explanation, and value type.

\clearpage
\clearpage
\begin{center}
\begin{figure}[H]
    \vspace*{0 cm}
    \hspace*{0cm}
    \begin{minipage}{\textwidth}
        \makebox[\textwidth]{
            \includegraphics[height=0.75\paperheight,keepaspectratio]{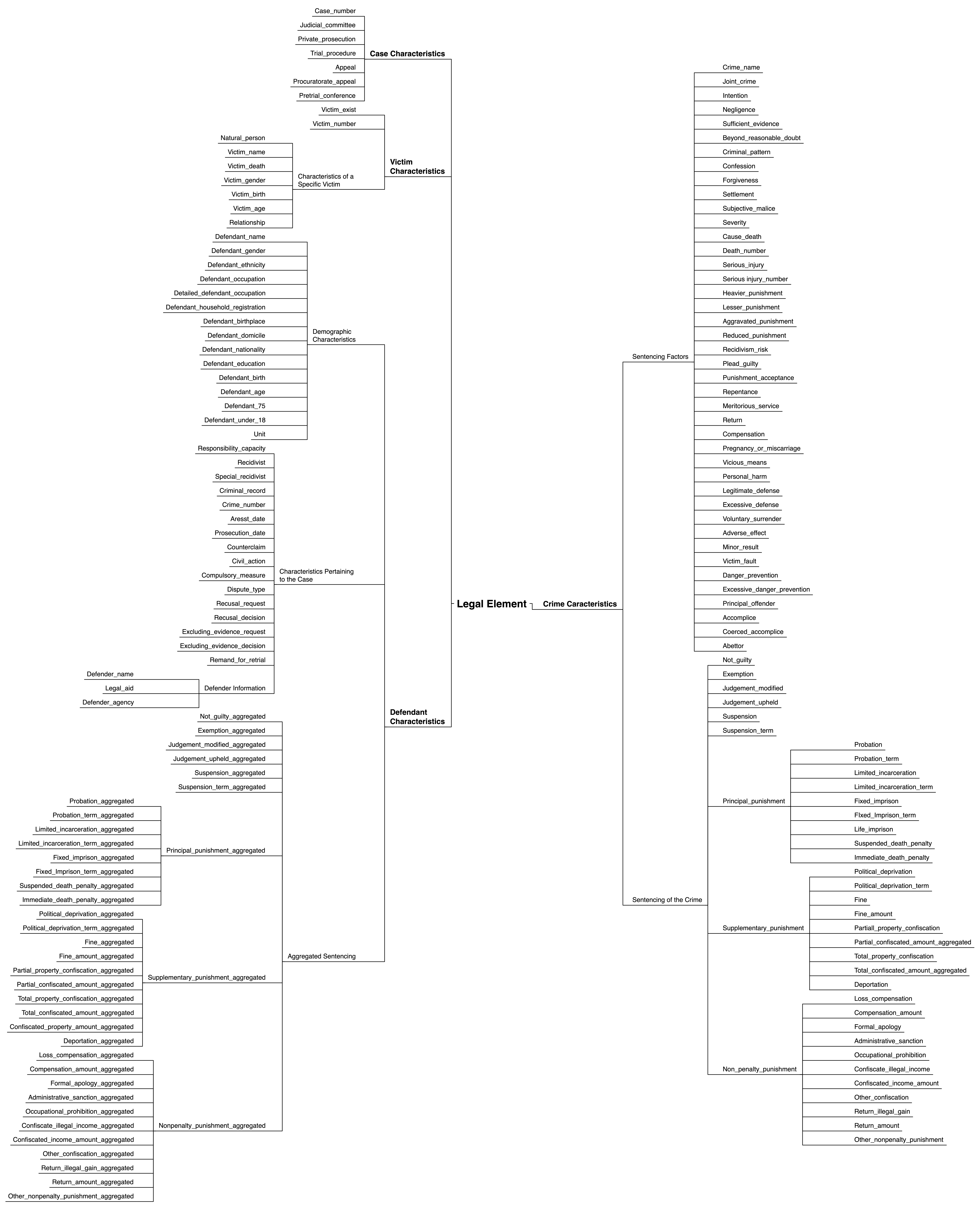}
        }
        \captionof{figure}{\centering The Detailed Element Schema of LEEC.}
    \end{minipage}
\end{figure}
\end{center}
\clearpage

\begin{table}
\centering
\begin{minipage}{\textwidth}
\newcolumntype{Y}{>{\hsize=.7\hsize\arraybackslash}X}
\newcolumntype{Z}{>{\hsize=1.45\hsize\arraybackslash}X}
\newcolumntype{S}{>{\hsize=0.85\hsize\arraybackslash}X}
\begin{tabularx}{\textwidth}{Y|Z|S}
\hline
\hline
\textbf{Element Name} & \textbf{Explanation} & \textbf{Value Type} \\
\hline
\multicolumn{3}{c}{\textbf{Case Characteristics}}\\
\hline
Case\_number & A CASE\_NUMBER element refers to the ID assigned by the court that uniquely identifies each judicial documents. & Extracted from the Specific Content of Judicial Documents \\
\hline
Judicial\_committee & A JUDICIAL\_COMMITTEE element refers to whether the court submitted the case to the judicial committee for discussion. & 1 = Yes, 0 = No\\
\hline
Private\_prosecution  & A PRIVATE\_PROSECUTION element refers to whether the victim, the victim's legal representative, or close relative institute an action directly in a people's court in a case of private prosecution.  & 1 = Yes, 0 = No, Non-mention is treated as 0 \\
\hline
Trial\_procedure  &  A TRIAL\_PROCEDURE element refers to the procedure applied to the trial of the case.  & Summary, Formal, Fast-Track Sentencing, Transfer from Summary to Formal Procedure, Transfer from Fast-Track to Formal Procedure\\
\hline
Appeal  &  AN APPEAL element refers to whether the defendant appealed the case.  & 1 = Yes, 0 = No\\
\hline
Procuratorate\_appeal  &  A PROCURATORATE\_APPEAL element refers to whether the people's procuratorate files an appeal to the people's court at the higher level.  & 1 = Yes, 0 = No\\
\hline
Pretrial\_conference  &  A PRETRIAL\_CONFERENCE element refers to whether the court determined that a pretrial conference for a case should be held.  & 1 = Yes, 0 = No\\
\hline

\multicolumn{3}{c}{\textbf{Victim Characteristics (I)}}\\
\hline

Victim\_exist  &  A VICTIM\_EXIST element refers to there was a victim whose legal rights and interests have been violated by a criminal act in a criminal case according to the court.  & 1 = Yes, 0 = No\\
\hline
Victim\_number  &  A VICTIM\_NUMBER element refers to the number of victims according to the information provided by the court. This element is annotated only when elements VICTIM\_EXIST is assigned a value of 1.  & Numeric Value Calculated Based on the Content of Judicial Documents\\
\hline
Natural\_person  &  A NATURAL\_PERSON element refers to whether the victim is a natural person in the biological sense. This element is annotated only when elements VICTIM\_EXIST is assigned a value of 1.  & 1 = Yes, 0 = No\\
\hline
Victim\_name  &  A VICTIM\_NAME element refers to the name of the victim according to the court. This element is annotated only when elements VICTIM\_EXIST and NATURAL\_PERSON are both assigned a value of 1.  & Extracted from the Specific Content of Judicial Documents\\
\hline
Victim\_death  &  A VICTIM\_DEATH element refers to whether a specific victim was dead.  This element is annotated only when elements VICTIM\_EXIST and NATURAL\_PERSON are both assigned a value of 1.  & 1 
 = Yes, 0 = No\\
\hline
Victim\_gender  &  A VICTIM\_GENDER element refers to the gender of the victim according to the court. This element is annotated only when elements VICTIM\_EXIST and NATURAL\_PERSON are both assigned a value of 1.  & Extracted from the Specific Content of Judicial Documents\\
\hline
Victim\_birth  &  A VICTIM\_BIRTH element refers to the date of birth of the victim according to the court. This element is annotated only when elements VICTIM\_EXIST and NATURAL\_PERSON are both assigned a value of 1.  & Extracted from the Specific Content of Judicial Documents\\
\hline
Victim\_age  &  A VICTIM\_AGE element refers to the age of the victim according to the court. This element is annotated only when elements VICTIM\_EXIST and NATURAL\_PERSON are both assigned a value of 1.  & Extracted from the Specific Content of Judicial Documents\\
\hline
\hline
\end{tabularx}

\captionof{table}{\centering List of detailed element information (I).}
\end{minipage}
\end{table}
\clearpage

\begin{table}
\centering
\begin{minipage}{\textwidth}
\newcolumntype{Y}{>{\hsize=.7\hsize\arraybackslash}X}
\newcolumntype{Z}{>{\hsize=1.45\hsize\arraybackslash}X}
\newcolumntype{S}{>{\hsize=0.85\hsize\arraybackslash}X}
\begin{tabularx}{\textwidth}{Y|Z|S}
\hline
\hline
\textbf{Element Name} & \textbf{Explanation} & \textbf{Value Type} \\
\hline
\multicolumn{3}{c}{\textbf{Victim Characteristics (II)}}\\
\hline
Relationship & A RELATIONSHIP element refers to the relationship between the defendant and the victim based on the content of judicial documents.  & Relationship between Non-natural Persons, Marital Relationship, Close Relatives, Other Relatives, Acquainted, Unknown\\
\hline
\multicolumn{3}{c}{\textbf{Defendant Characteristics (I)}}\\
\hline
Defendant\_name  &  A DEFENDANT\_NAME element refers to the name of the defendant.  & Extracted from the Specific Content of Judicial Documents\\
\hline
Defendant\_gender  &  A DEFENDANT\_GENDER element refers to the gender of the defendant.  & 1 = Male; 0 = Female\\
\hline
Defendant\_ethnicity  &  A DEFENDANT\_ETHNICITY element refers to the ethnicity of the defendant.  & Extracted from the Specific Content of Judicial Documents\\
\hline
Defendant\_occupation  &  A DEFENDANT\_OCCUPATION element refers to the occupation of the defendant categorized into four types.  & Extracted from the Specific Content of Judicial Documents\\
\hline
\url{Detailed\_defendant\_occupation}  &  A DETAILED\_DEFENDANT\_OCCUPATION element refers to the detailed occupation of the defendant as explicitly stated in the judicial document.  & Extracted from the Specific Content of Judicial Documents\\
\hline
\url{Defendant\_household\_registration}  &  A DEFENDANT\_HOUSEHOLD\_REGISTRATION element refers to the place of registered permanent residence of the defendant, also known as \textit{Hukou} in Chinese.  & Extracted from the Specific Content of Judicial Documents\\
\hline
Defendant\_birthplace  &  A DEFENDANT\_BIRTHPLACE element refers to the place of birth of the defendant.  & Extracted from the Specific Content of Judicial Documents\\
\hline
Defendant\_domicile  &  A DEFENDANT\_DOMICILE element refers to the address of the defendant.  & Extracted from the Specific Content of Judicial Documents\\
\hline
Defendant\_nationality  &  A DEFENDANT\_NATIONALITY element refers to the nationality of the defendant.  & Extracted from the Specific Content of Judicial Documents\\
\hline
Defendant\_education  &  A DEFENDANT\_EDUCATION element refers to to the level of formal education or academic degree attained by the defendant.  & Illiteracy, Under Secondary School, Secondary School, Primary School, Regular Senior High School, Secondary Vocational School, Tertiary Vocational School, Bachelor's Degree, Master's Degree, Doctor's Degree\\
\hline
Defendant\_birth  &  A DEFENDANT\_BIRTH element refers to the date of birth of the defendant.  & Extracted from the Specific Content of Judicial Documents\\
\hline
Defendant\_age  &  A DEFENDANT\_AGE element refers to the age of the defendant.  & Extracted from the Specific Content of Judicial Documents\\
\hline
Defendant\_75  &  A DEFENDANT\_75 element refers to whether the court determined that the defendant is above the age of 75, which is a mitigating circumstance in Chinese criminal law, and thereby may be explicitly stated in the judicial document.  & 1 = Yes; 0 = No\\
\hline
Defendant\_under\_18 &  A DEFENDANT\_UNDER\_18 element refers to whether the court determined that the defendant is under the age of 18, which is a mitigating circumstance in Chinese criminal law, and thereby may be explicitly stated in the judicial document.   & 1 = Yes; 0 = No\\
\hline
Unit &  A UNIT element refers to whether the defendant is a company, enterprise, institution, organization, or group. & 1 = Yes; 0 = No\\
\hline
Responsibility\_capacity &  A RESPONSIBILITY\_CAPACITY element refers to  the court determined that the level of the defendant's ability to take responsibility for crimes. & Full Criminal Responsibility Capacity  Criminal Responsibility Incapacity  Relatively Criminal Responsibility Incapacity  Partial Criminal Responsibility Capacity\\
\hline
Recidivist &  A RECIDIVIST element refers to whether the court determined that the defendant was a recidivist. & 1 = Yes; 0 = No\\
\hline
\hline
\end{tabularx}

\captionof{table}{\centering List of detailed element information (II).}
\end{minipage}
\end{table}
\clearpage

\begin{table}
\centering
\begin{minipage}{\textwidth}
\newcolumntype{Y}{>{\hsize=.7\hsize\arraybackslash}X}
\newcolumntype{Z}{>{\hsize=1.45\hsize\arraybackslash}X}
\newcolumntype{S}{>{\hsize=0.85\hsize\arraybackslash}X}
\begin{tabularx}{\textwidth}{Y|Z|S}
\hline
\hline
\textbf{Element Name} & \textbf{Explanation} & \textbf{Value Type} \\
\hline
\multicolumn{3}{c}{\textbf{Defendant Characteristics (II)}}\\
\hline
Special\_recidivist &  A SPECIAL\_RECIDIVIST element refers to whether the court determined that the defendant was a special recidivist as stipulated in the Chinese criminal law. & 1 = Yes; 0 = No\\
\hline
Criminal\_record &  A CRIMINAL\_RECORD element refers to whether the court determined that the defendant had a previous criminal record. & 1 = Yes; 0 = No\\
\hline
Crime\_number &  A CRIME\_NUMBER element refers to the total number of the crime name of a specific defendant. & Numeric Value Calculated Based on the Content of Judicial Documents\\
\hline
Arrest\_date &  An ARREST\_DATE element refers to the date of the execution of arrest by criminal justice authorities. & Extracted from the Specific Content of Judicial Documents\\
\hline
Prosecution\_date &  A PROSECUTION\_DATE element refers to the date of the initiation of a public prosecution by the people's procuratorate. & Extracted from the Specific Content of Judicial Documents\\
\hline
Counterclaim &  A COUNTERCLAIM element refers to whether the defendant files a counterclaim against the private prosecutor in a private criminal prosecution case or the victim in the incidental civil action part of the criminal incidental civil case. & 1 = Yes; 0 = No; Non-mention is treated as 0\\
\hline
Civil\_action &  An CIVIL\_ACTION element refers to whether the court determined that the incidental civil action is instituted. & 1 = Yes; 0 = No; Non-mention is treated as 0\\
\hline
Compulsory\_measure &  A COMPULSORY\_MEASURE element refers to the methods to restrict a certain degree of personal freedom for criminal suspects and defendants by criminal justice authorities. & Custody/Forced Appearance/Granted Bail/Residential Confinement/Detention/Arrest\\
\hline
Dispute\_type &  A DISPUTE\_TYPE element refers to the type of dispute involved in the case, including disputes with clearly identified victims, disputes between neighbors, family disputes, and disputes without clearly identified victims. & Family Disputes / Disputes among Neighbors / Other Disputes with  Victims / Disputes without Victims\\
\hline
Recusal\_request &  A RECUSAL\_REQUEST element refers to whether the parties requested for recusal. & 1 = Yes; 0 = No\\
\hline
Recusal\_decision &  A RECUSAL\_DECISION element refers to whether the court approved of a recusal request. & 1 = Yes; 0 = No\\
\hline
Excluding\_evidence\_request &  AN  EXCLUDING\_EVIDENCE\_APPLICATION element refers to whether parties and their defenders or litigation representatives requested for excluding illegal evidence. & 1 = Yes; 0 = No\\
\hline
Excluding\_evidence\_decision &  AN  EXCLUDING\_EVIDENCE\_DECISION element refers to whether the court approved of a  request for excluding illegal evidence. & 1 = Yes; 0 = No\\
\hline
Remand\_for\_retrial &  A REMAND\_FOR\_RETRIAL element refers to whether the court determined that the case shall be remanded for retrial. & 1 = Yes; 0 = No\\
\hline
Defender\_name &  A DEFENDER\_NAME element refers to the name of the defender. & Extracted from the Specific Content of Judicial Documents\\
\hline
Legal\_aid &  A LEGAL\_AID element refers to whether the defender was designated by the legal aid agency. & 1 = Yes; 0 = No\\
\hline
Defender\_agency &  A DEFENDER\_AGENCY element refers to the agency of the defender. & Extracted from the Specific Content of Judicial Documents\\
\hline
Not\_guilty\_aggregated &  A NOT\_GUILTY\_AGGREGATED element refers to whether the court determined that the defendant was not guilty. & 1 = Yes; 0 = No; Non-mention is treated as 0\\
\hline
Exemption\_aggregated &  An EXEMPTION\_AGGREGATED element refers to whether the defendant is exempted from criminal punishment as determined by the court. & 1 = Yes; 0 = No; Non-mention is treated as 0\\
\hline

\hline
\end{tabularx}

\captionof{table}{\centering List of detailed element information (III).}
\end{minipage}
\end{table}
\clearpage

\begin{table}
\centering
\begin{minipage}{\textwidth}
\newcolumntype{Y}{>{\hsize=.7\hsize\arraybackslash}X}
\newcolumntype{Z}{>{\hsize=1.45\hsize\arraybackslash}X}
\newcolumntype{S}{>{\hsize=0.85\hsize\arraybackslash}X}
\begin{tabularx}{\textwidth}{Y|Z|S}
\hline
\hline
\textbf{Element Name} & \textbf{Explanation} & \textbf{Value Type} \\
\hline
\multicolumn{3}{c}{\textbf{Defendant Characteristics (III)}}\\
\hline
\url{Judgement\_modified\_aggregated} &  A JUDGEMENT\_MODIFIED\_AGGREGATED element refers to whether the court determined that full or partial revision of sentence is made in the aggregated sentencing of a defendant. & 1 = Yes; 0 = No; Non-mention is treated as 0\\
\hline
\url{Judgement\_upheld\_aggregated} &  A JUDGEMENT\_UPHELD\_AGGREGATED element refers to whether the court decided to uphold the previous judgment in the aggregated sentencing of a defendant. & 1 = Yes; 0 = No; Non-mention is treated as 0\\
\hline
\url{Suspension\_aggregated} & A SUSPENSION\_AGGREGATED element refers to whether the court determined a suspension of sentence in the aggregated sentencing of a defendant. & 1 = Yes; 0 = No; Non-mention is treated as 0\\
\hline
\url{Suspension\_term\_aggregated} & A SUSPENSION\_TERM\_AGGREGATED element refers to the probation period for suspension as determined by the court in the aggregated sentencing of a defendant. This element is annotated only when element SUSPENSION\_AGGREGATED is assigned a value of 1. & Extracted from the Specific Content of Judicial Documents\\
\hline
\url{Probation_aggregated} &  A PROBATION\_AGGREGATED element refers to whether the defendant was subject to probation as determined by the court in the aggregated sentencing. &  1 = Yes; 0 = No; Non-mention is treated as 0\\
\hline
\url{Probation\_term\_aggregated} &  A PROBATION\_TERM\_AGGREGATED element refers to the term of probation as determined by the court in the aggregated sentencing of a defendant. This element is annotated only when element PROBATION\_AGGREGATED is assigned a value of 1. & Extracted from the Specific Content of Judicial Documents\\
\hline

\url{Limited_incarceration_aggregated} &  A LIMITED\_INCARNATION\_AGGREGATED element refers to whether the defendant was subject to limited incarceration as determined by the court in the aggregated sentencing. &  1 = Yes; 0 = No; Non-mention is treated as 0\\
\hline
\url{Limited\_incarceration\_term\_aggregated} &  A LIMITED\_INCARCERATION\_TERM\_AGGREGATED element refers to the term of limited incarceration as determined by the court in the aggregated sentencing of a defendant. This element is annotated only when element LIMITED\_INCARNATION\_AGGREGATED is assigned a value of 1. & Extracted from the Specific Content of Judicial Documents\\
\hline

\url{Fixed_imprison_aggregated} &  A FIXED\_IMPRISON\_AGGREGATED element refers to whether the defendant was subject to fixed incarceration as determined by the court in the aggregated sentencing. &  1 = Yes; 0 = No; Non-mention is treated as 0\\
\hline
\url{Fixed\_imprison\_term\_aggregated} &  A FIXED\_IMPRISON\_TERM\_AGGREGATED element refers to the term of fixed-term imprisonment as determined by the court in the aggregated sentencing of a defendant. This element is annotated only when element FIXED\_IMPRISON\_AGGREGATED is assigned a value of 1. & Extracted from the Specific Content of Judicial Documents\\
\hline

\url{Life_imprison_aggregated} &  A LIFE\_IMPRISON\_AGGREGATED element refers to whether the defendant was subject to life imprisonment as determined by the court in the aggregated sentencing. &  1 = Yes; 0 = No; Non-mention is treated as 0\\
\hline

\url{Suspended_death_penalty_aggregated} &  A SUSPENDED\_DEATH\_PENALTY\_AGGREGATED element refers to whether the defendant was subject to a suspended death penalty as determined by the court in the aggregated sentencing. &  1 = Yes; 0 = No; Non-mention is treated as 0\\
\hline
\url{Immediate_death_penalty_aggregated} &  A IMMEDIATE\_DEATH\_PENALTY\_AGGREGATED element refers to whether the defendant was subject to an immediate death penalty as determined by the court in the aggregated sentencing. &  1 = Yes; 0 = No; Non-mention is treated as 0\\
\hline

\hline
\hline
\end{tabularx}

\captionof{table}{\centering List of detailed element information (IV).}
\end{minipage}
\end{table}
\clearpage

\begin{table}
\centering
\begin{minipage}{\textwidth}
\newcolumntype{Y}{>{\hsize=.7\hsize\arraybackslash}X}
\newcolumntype{Z}{>{\hsize=1.45\hsize\arraybackslash}X}
\newcolumntype{S}{>{\hsize=0.85\hsize\arraybackslash}X}
\begin{tabularx}{\textwidth}{Y|Z|S}
\hline
\hline
\textbf{Element Name} & \textbf{Explanation} & \textbf{Value Type} \\
\hline
\multicolumn{3}{c}{\textbf{Defendant Characteristics (IV)}}\\
\hline
\url{Political\_deprivation\_aggregated} &  A POLITICAL\_DEPRIVATION\_AGGREGATED element refers to whether the defendant was subject to the deprivation of political rights as determined by the court in the aggregated sentencing. &  1 = Yes; 0 = No; Non-mention is treated as 0\\
\hline
\url{Political\_deprivation\_term\_aggregated} &  A POLITICAL\_DEPRIVATION\_TERM\_AGGREGATED element refers to the term of deprivation of political rights as determined by the court in the aggregated sentencing of a defendant. This element is annotated only when element POLITICAL\_DEPRIVATION\_AGGREGATED is assigned a value of 1. & Extracted from the Specific Content of Judicial Documents\\
\hline

\url{Fine\_aggregated} &  A FINE\_AGGREGATED element refers to whether the defendant was fined as determined by the court in the aggregated sentencing. &  1 = Yes; 0 = No; Non-mention is treated as 0\\
\hline
\url{Fine\_amount\_aggregated} & A FINE\_AMOUNT\_AGGREGATED element refers to the amount of the fine as determined by the court in the aggregated sentencing of a defendant. This element is annotated only when element FINE\_AGGREGATED is assigned a value of 1. & Extracted from the Specific Content of Judicial Documents\\
\hline
\url{Partial\_property\_confiscation\_aggregated} &  A PARTIAL\_PROPERTY\_CONFISCATION\_AGGREGATED element refers to whether the defendant was subject to confiscation of a part of his or her property as determined by the court in the aggregated sentencing. &  1 = Yes; 0 = No; Non-mention is treated as 0\\
\hline
\url{Total\_property\_confiscation\_aggregated} &  A TOTAL\_PROPERTY\_CONFISCATION\_AGGREGATED element refers to whether the defendant was subject to confiscation of all of his or her property as determined by the court in the aggregated sentencing. &  1 = Yes; 0 = No; Non-mention is treated as 0\\
\hline
\url{Confiscated\_property\_amount\_aggregated} & A CONFISCATED\_PROPERTY\_AMOUNT\_AGGREGATED element refers to the amount of confiscated property as determined by the court in the aggregated sentencing of a defendant. This element is annotated only when either element PARTIAL\_PROPERTY\_CONFISCATION\_AGGREGATED or element TOTAL\_PROPERTY\_CONFISCATION\_AGGREGATED is assigned a value of 1. & Extracted from the Specific Content of Judicial Documents\\
\hline
\url{Deportation\_aggregated} &  A DEPORTATION\_AGGREGATED element refers to whether the defendant was subject to deportation as determined by the court in the aggregated sentencing. &  1 = Yes; 0 = No; Non-mention is treated as 0\\
\hline

\url{Loss\_compensation\_aggregated} & A LOSS\_COMPENSATION\_AGGREGATED element refers to whether the defendant was ordered to make compensation for the economic loss as determined by the court in the aggregated sentencing. &  1 = Yes; 0 = No; Non-mention is treated as 0\\
\hline
\url{Compensation\_amount\_aggregated} & A COMPENSATION\_AMOUNT\_AGGREGATED element refers to the amount of making compensation for the economic loss as determined by the court in the aggregated sentencing of a defendant. This element is annotated only when element LOSS\_COMPENSATION\_AGGREGATED is assigned a value of 1. & Extracted from the Specific Content of Judicial Documents\\
\hline

\url{Formal\_apology\_aggregated} & A FORMAL\_APOLOGY\_AGGREGATED element refers to whether the defendant was ordered to make a statement of repentance or formal apology as determined by the court in the aggregated sentencing. &  1 = Yes; 0 = No; Non-mention is treated as 0\\
\hline
\url{Administrative\_sanction\_aggregated} & An ADMINISTRATIVE\_SANCTION\_AGGREGATED element refers to whether the defendant was subjected to administrative sanctions by the relevant department as determined by the court in the aggregated sentencing. &  1 = Yes; 0 = No; Non-mention is treated as 0\\

\hline
\hline
\end{tabularx}

\captionof{table}{\centering List of detailed element information (V).}
\end{minipage}
\end{table}
\clearpage

\begin{table}
\centering
\begin{minipage}{\textwidth}
\newcolumntype{Y}{>{\hsize=.7\hsize\arraybackslash}X}
\newcolumntype{Z}{>{\hsize=1.45\hsize\arraybackslash}X}
\newcolumntype{S}{>{\hsize=0.85\hsize\arraybackslash}X}
\begin{tabularx}{\textwidth}{Y|Z|S}
\hline
\hline
\textbf{Element Name} & \textbf{Explanation} & \textbf{Value Type} \\
\hline
\multicolumn{3}{c}{\textbf{Defendant Characteristics (V)}}\\
\hline
\url{Occupational\_prohibition\_aggregated} & An OCCUPATIONAL\_PROHIBITION\_AGGREGATED element refers to whether the defendant was subjected to occupational prohibition as determined by the court in the aggregated sentencing. &  1 = Yes; 0 = No; Non-mention is treated as 0\\
\hline

\url{Confiscate\_illegal\_income\_aggregated} & A CONFISCATE\_ILLEGAL\_INCOME\_AGGREGATED element refers to whether the defendant was subjected to the confiscation of illegal income as determined by the court in the aggregated sentencing. &  1 = Yes; 0 = No; Non-mention is treated as 0\\
\hline
\url{Confiscated\_income\_amount\_aggregated} & A CONFISCATED\_INCOME\_AMOUNT\_AGGREGATED element refers to the amount of confiscated income as determined by the court in the aggregated sentencing of a defendant. This element is annotated only when element CONFISCATE\_ILLEGAL\_INCOME\_AGGREGATED is assigned a value of 1. & Extracted from the Specific Content of Judicial Documents\\
\hline

\url{Other\_confiscation\_aggregated} & An OTHER\_CONFISCATION\_AGGREGATED element refers to whether the defendant was subjected to the confiscation of other objects related to the crime as determined by the court in the aggregated sentencing. &  1 = Yes; 0 = No; Non-mention is treated as 0\\
\hline

\url{Return\_illegal\_gain\_aggregated} & A RETURN\_ILLEGAL\_GAIN\_AGGREGATED element refers to the order for returning illegal gains and compensations as determined by the court in the aggregated sentencing of a defendant. & 1 = Yes; 0 = No; Non-mention is treated as 0\\
\hline
\url{Return\_amount\_aggregated} & A RETURN\_AMOUNT\_AGGREGATED element refers to the amount of returning illegal gains and compensations as ordered by the court in the aggregated sentencing of a defendant. This element is annotated only when element RETURN\_ILLEGAL\_GAIN\_AGGREGATED is assigned a value of 1. & Extracted from the Specific Content of Judicial Documents\\
\hline
\url{Other\_nonpenalty\_punishment\_aggregated} & A OTHER\_NONPENALTY\_PUNISHMENT\_AGGREGATED element refers to the order of other types of nonpenalty punishment as determined by the court in the aggregated sentencing of a defendant. & 1 = Yes; 0 = No; Non-mention is treated as 0\\
\hline
\multicolumn{3}{c}{\textbf{Crime Characteristics (I)}}\\
\hline
Crime\_name  &  A CRIME\_NAME element refers to the charges against the defendant.  & Extracted from the Specific Content of Judicial Documents\\
\hline
Joint\_crime  &  A JOINT\_CRIME element refers to whether the court determined that an intentional crime was committed by two or more persons or units jointly.  & 1 = Yes; 0 = No\\
\hline
Intention  &  An INTENTION element refers to the subjective aspect of the defendant in an intentional crime as determined by the court. & Direct Intention/Indirect Intention\\
\hline
Negligence  &  A NEGLIGENCE element refers to the subjective aspect of the criminal in a negligent crime as determined by the court. & Carelessness/Overconfidence\\
\hline
Sufficient\_evidence   &  A SUFFICIENT\_EVIDENCE element refers to whether the court determined that evidence is sufficient. & 1 = Yes; 0 = No\\
\hline
Beyond\_reasonable\_doubt  &  A BEYOND\_REASONABLE\_DOUBT element refers to whether the court determined that the evidence is strong enough to rule out any reasonable doubt. & 1 = Yes; 0 = No\\
\hline
Criminal\_pattern &  A CRIMINAL\_PATTERN element refers to the pattern of the crime. & Preparation/Discontinuation/Criminal Attempt/Consummation\\
\hline
Confession  &  A CONFESSION element refers to whether the court determined that the defendant confessed to authorities. & 1 = Yes; 0 = No\\
\hline
Forgiveness  &  A FORGIVENESS element refers to whether the court determined that the defendant had obtained forgiveness from victims or their close relatives. & 1 = Yes; 0 = No\\

\hline
\hline
\end{tabularx}

\captionof{table}{\centering List of detailed element information (VI).}
\end{minipage}
\end{table}
\clearpage

\begin{table}
\centering
\begin{minipage}{\textwidth}
\newcolumntype{Y}{>{\hsize=.7\hsize\arraybackslash}X}
\newcolumntype{Z}{>{\hsize=1.45\hsize\arraybackslash}X}
\newcolumntype{S}{>{\hsize=0.85\hsize\arraybackslash}X}
\begin{tabularx}{\textwidth}{Y|Z|S}
\hline
\hline
\textbf{Element Name} & \textbf{Explanation} & \textbf{Value Type} \\
\hline
\multicolumn{3}{c}{\textbf{Crime Characteristics (II)}}\\
\hline
Settlement  &  A SETTLEMENT element refers to whether both parties reach a settlement. & 1 = Yes; 0 = No\\
\hline
Subjective\_malice  &  A SUBJECTIVE\_MALICE element referes to the degree of the subjective malice of the defendant as determined by the court. & No or Low Subjective Malice/High Subjective Malice\\
\hline
Severity  &  A SEVERITY element refers to the level of severity of the crime as determined by the court. & Circumstances Clearly Minor/Circumstances Minor/Circumstances Serious or Execrable/Circumstances Very Serious or Execrable\\
\hline
Cause\_death  &  A CAUSE\_DEATH element refers to whether the court determined that whether any death occurred because of the crime. & 1 = Yes; 0 = No\\
\hline
Death\_number  &  A DEATH\_NUMBER element refers to the total number of victims who died because of the crime according to the court. This element is annotated only when element VICTIM\_DEATH is assigned a value of 1. & Numeric Value Calculated Based on the Content of Judicial Documents\\
\hline
Serious\_injury  &  A SERIOUS\_INJURY element refers to whether the court determined that serious injury of any victim is caused by the crime. & 1 = Yes; 0 = No\\
\hline
Serious\_injury\_number  &  A SERIOUS\_INJURY\_NUMBER element refers to the total number of victims who were seriously injured by the crime according to the court. This element is annotated only when element SERIOUS\_INJURY is assigned a value of 1. & Numeric Value Calculated Based on the Content of Judicial Documents\\
\hline
Heavier\_punishment  &  A HEAVIER\_PUNISHMENT element refers to the number of circumstances leading to heavier punishment within the legally prescribed limits of punishment according to the court. & Numeric Value Calculated Based on the Content of Judicial Documents\\
\hline
Lesser\_punishment  &  A LESSER\_PUNISHMENT element refers to the number of circumstances leading to lesser punishment within the legally prescribed limits of punishment according to the court. & Numeric Value Calculated Based on the Content of Judicial Documents\\
\hline
Aggravated\_punishment  &  An AGGRAVATED\_PUNISHMENT element refers to the number of circumstances leading to aggravated punishment above the legally prescribed limits of punishment according to the court. & Numeric Value Calculated Based on the Content of Judicial Documents\\
\hline
Mitigated\_punishment  &  A MITIGATED\_PUNISHMENT element refers to the number of circumstances leading to reduced punishment below the legally prescribed limits of punishment according to the court. & Numeric Value Calculated Based on the Content of Judicial Documents\\
\hline
Reoffending\_danger  & A REOFFENDING\_DANGER element refers to whether the court determined that the defendant would likely to commit any crime again. & 1 = Yes; 0 = No\\
\hline
Plead\_guilty  & A PLEAD\_GUILTY element refers to whether the court determined that the defendant pled guilty. & 1 = Yes; 0 = No\\
\hline
Punishment\_acceptance  & A PUNISHMENT\_ACCEPTANCE element refers to whether the court determined that the defendant accepted punishment. & 1 = Yes; 0 = No\\
\hline
Repentance  & A REPENTANCE element refers to whether the court determined that the defendant had showed genuine repentance. & 1 = Yes; 0 = No\\
\hline
Meritorious\_service  & A MERITORIOUS\_SERVICE element refers to whether the court determined that the defendant performed meritorious service. & 1 = Yes; 0 = No\\
\hline
Return  & A RETURN element refers to whether the court determined that the defendant actively returned the property that the defendant acquired illegally or equivalent amount of money. & 1 = Yes; 0 = No\\
\hline
Compensation  & A COMPENSATION element refers to whether the court determined that the defendant actively made compensation to victims actively. & 1 = Yes; 0 = No\\

\hline
\hline
\end{tabularx}

\captionof{table}{\centering List of detailed element information (VII).}
\end{minipage}
\end{table}
\clearpage

\begin{table}
\centering
\begin{minipage}{\textwidth}
\newcolumntype{Y}{>{\hsize=.7\hsize\arraybackslash}X}
\newcolumntype{Z}{>{\hsize=1.45\hsize\arraybackslash}X}
\newcolumntype{S}{>{\hsize=0.85\hsize\arraybackslash}X}
\begin{tabularx}{\textwidth}{Y|Z|S}
\hline
\hline
\textbf{Element Name} & \textbf{Explanation} & \textbf{Value Type} \\
\hline
\multicolumn{3}{c}{\textbf{Crime Characteristics (III)}}\\
\hline
Pregnancy\_or\_miscarriage  & A PREGNANCY\_OR\_MISCARRIAGE element refers to whether the court determined that the defendant was pregnant or had a miscarriage during prosecution or trial. & 1 = Yes; 0 = No\\
\hline
Vicious\_means  & A VICIOUS\_MEANS element refers to whether the court determined that the crime was conducted using vicious means. & 1 = Yes; 0 = No\\
\hline
Personal\_harm & A PERSONAL\_HARM element refers to whether the court determined that the defendant posed a high risk of or caused personal harm. & 1 = Yes; 0 = No\\
\hline
Legitimate\_defense  & A LEGITIMATE\_DEFENSE element refers to whether whether the court determined that the act of the defendant was legitimate defense. & 1 = Yes; 0 = No\\
\hline
Excessive\_defense  & An EXCESSIVE\_DEFENSE element refers to whether the court determined that defendant's defense noticeably exceeded the necessary limits. & 1 = Yes; 0 = No\\
\hline
Voluntary\_Surrender  & A VOLUNTARY\_SURRENDER element refers to whether the court determined that the defendant voluntarily surrendered to the police and gave a true account of one's crime after committing it. & 1 = Yes; 0 = No\\
\hline
Adverse\_effect  & An ADVERSE\_EFFECT element refers to whether the court determined that the social effects caused by the defendant were significantly adverse. & 1 = Yes; 0 = No\\
\hline
Minor\_result  & A MINOR\_RESULT element refers to whether the court determined that the circumstances of the alleged conduct are obviously minor, causing no serious harm. & 1 = Yes; 0 = No\\
\hline
Victim\_fault  & A VICTIM\_FAULT element refers to whether the court determined that victim was in fault for the crime. & 1 = Yes; 0 = No\\
\hline
Danger\_prevention  & An DANGER\_PREVENTION element refers to whether the court determined that the defendant's act was a legitimate prevention of urgent danger. & 1 = Yes; 0 = No\\
\hline
Excessive\_danger\_prevention  & An EXCESSIVE\_DANGER\_PREVENTION element refers to  whether the court determined that the defendant's act was a prevention of urgent danger that exceeded the necessary limits and caused undue harm.  & 1 = Yes; 0 = No\\
\hline
Principal\_offender  & A PRINCIPAL\_OFFENDER element refers to whether the court determined that the defendant organized and leads a criminal group in conducting criminal activities or played a principal role in a joint crime. This element is annotated only when JOINT\_CRIME element is assigned a value of 1. & 1 = Yes; 0 = No\\
\hline
Accomplice  & An ACCOMPLICE element refers to whether the court determined that the defendant played a secondary or supplementary role in a joint crime. This element is annotated only when JOINT\_CRIME element is assigned a value of 1. & 1 = Yes; 0 = No\\
\hline
Coerced\_accomplice & A COERCED\_ACCOMPLICE element refers to whether the court determined that the defendant was coerced to participate in a crime shall. This element is annotated only when JOINT\_CRIME element is assigned a value of 1. & 1 = Yes; 0 = No\\
\hline
Abettor & An ABETTOR element refer to whether the court determined that the defendant was the one who instigated others to commit a crime and should be punished according to the role he played in the joint crime. This element is annotated only when JOINT\_CRIME element is assigned a value of 1. & 1 = Yes; 0 = No\\
\hline
Not\_guilty &  A NOT\_GUILTY element refers to whether the court determined that the defendant was not guilty of a specific crime. & 1 = Yes; 0 = No; Non-mention is treated as 0\\
\hline

\hline
\end{tabularx}

\captionof{table}{\centering List of detailed element information (VIII).}
\end{minipage}
\end{table}
\clearpage

\begin{table}
\centering
\begin{minipage}{\textwidth}
\newcolumntype{Y}{>{\hsize=.7\hsize\arraybackslash}X}
\newcolumntype{Z}{>{\hsize=1.45\hsize\arraybackslash}X}
\newcolumntype{S}{>{\hsize=0.85\hsize\arraybackslash}X}
\begin{tabularx}{\textwidth}{Y|Z|S}
\hline
\hline
\textbf{Element Name} & \textbf{Explanation} & \textbf{Value Type} \\
\hline
\multicolumn{3}{c}{\textbf{Crime Characteristics (IV)}}\\
\hline

Exemption &  An EXEMPTION element refers to whether the defendant is exempted from criminal punishment of a specific crime as determined by the court. & 1 = Yes; 0 = No; Non-mention is treated as 0\\
\hline
\url{Judgement\_modified} &  A\_JUDGEMENT\_MODIFIED element refers to whether the court determined that full revision and partial reversal with partial revision of sentence should be made of a specific crime. & 1 = Yes; 0 = No; Non-mention is treated as 0\\
\hline
\url{Judgement\_upheld} &  A JUDGEMENT\_UPHELD element refers to whether the court decided to uphold the previous judgment of a specific crime. & 1 = Yes; 0 = No; Non-mention is treated as 0\\
\hline

\url{Suspension} & A SUSPENSION element refers to whether the court determined a suspension of sentence for a specific crime. & 1 = Yes; 0 = No\\
\hline
\url{Suspension\_term} & A SUSPENSION\_TERM element refers to the probation period for suspension for a specific crime as determined by the court. This element is annotated only when element SUSPENSION is assigned a value of 1. & Extracted from the Specific Content of Judicial Documents\\
\hline
\url{Probation} &  A PROBATION element refers to whether the defendant was subject to probation for a specific crime as determined by the court. &  1 = Yes; 0 = No; Non-mention is treated as 0\\
\hline
\url{Probation\_term} &  A PROBATION\_TERM element refers to the term of probation of a specific crime as determined by the court. This element is annotated only when element PROBATION is assigned a value of 1. & Extracted from the Specific Content of Judicial Documents\\
\hline

\url{Limited_incarceration} &  A LIMITED\_INCARNATION element refers to whether the defendant was subject to limited incarceration for a specific crime as determined by the court. &  1 = Yes; 0 = No; Non-mention is treated as 0\\
\hline
\url{Limited\_incarceration\_term} &  A LIMITED\_INCARCERATION\_TERM element refers to the term of limited incarceration as determined by the court for a specific crime of a defendant. This element is annotated only when element LIMITED\_INCARNATION is assigned a value of 1. & Extracted from the Specific Content of Judicial Documents\\
\hline

\url{Fixed_imprison} &  A FIXED\_IMPRISON element refers to whether the defendant was subject to fixed incarceration for a specific crime as determined by the court. &  1 = Yes; 0 = No; Non-mention is treated as 0\\
\hline
\url{Fixed\_imprison\_term} &  A FIXED\_IMPRISON\_TERM element refers to the term of fixed-term imprisonment as determined by the court for a specific crime of a defendant. This element is annotated only when element FIXED\_IMPRISON is assigned a value of 1. & Extracted from the Specific Content of Judicial Documents\\
\hline

\url{Life_imprison} &  A LIFE\_IMPRISON element refers to whether the defendant was subject to life imprisonment for a specific crime as determined by the court. &  1 = Yes; 0 = No; Non-mention is treated as 0\\
\hline

\url{Suspended_death_penalty} &  A SUSPENDED\_DEATH\_PENALTY element refers to whether the defendant was subject to a suspended death penalty for a specific crime as determined by the court. &  1 = Yes; 0 = No; Non-mention is treated as 0\\
\hline
\url{Immediate_death_penalty} &  A IMMEDIATE\_DEATH\_PENALTY element refers to whether the defendant was subject to an immediate death penalty for a specific crime as determined by the court. &  1 = Yes; 0 = No; Non-mention is treated as 0\\
\hline

\url{Political\_deprivation} &  A POLITICAL\_DEPRIVATION element refers to whether the defendant was subject to the deprivation of political rights for a specific crime as determined by the court. &  1 = Yes; 0 = No; Non-mention is treated as 0\\
\hline
\url{Political\_deprivation\_term} &  A POLITICAL\_DEPRIVATION\_TERM element refers to the term of deprivation of political rights for a specific crime of a defendant as determined by the court. This element is annotated only when element POLITICAL\_DEPRIVATION is assigned a value of 1. & Extracted from the Specific Content of Judicial Documents\\
\hline
\hline
\end{tabularx}

\captionof{table}{\centering List of detailed element information (IX).}
\end{minipage}
\end{table}
\clearpage

\begin{table}
\centering
\begin{minipage}{\textwidth}
\newcolumntype{Y}{>{\hsize=.7\hsize\arraybackslash}X}
\newcolumntype{Z}{>{\hsize=1.45\hsize\arraybackslash}X}
\newcolumntype{S}{>{\hsize=0.85\hsize\arraybackslash}X}
\begin{tabularx}{\textwidth}{Y|Z|S}
\hline
\hline
\textbf{Element Name} & \textbf{Explanation} & \textbf{Value Type} \\
\hline
\multicolumn{3}{c}{\textbf{Crime Characteristics (V)}}\\
\hline

\url{Fine} &  A FINE element refers to whether the defendant was fined for a specific crime as determined by the court. &  1 = Yes; 0 = No; Non-mention is treated as 0\\
\hline
\url{Fine\_amount} & A FINE\_AMOUNT element refers to the amount of the fine of a specific crime as determined by the court. This element is annotated only when element FINE is assigned a value of 1. & Extracted from the Specific Content of Judicial Documents\\
\hline
\url{Partial\_property\_confiscation} &  A PARTIAL\_PROPERTY\_CONFISCATION element refers to whether the defendant was subject to confiscation of a part of his or her property for a specific crime as determined by the court. &  1 = Yes; 0 = No; Non-mention is treated as 0\\
\hline
\url{Total\_property\_confiscation} &  A TOTAL\_PROPERTY\_CONFISCATION element refers to whether the defendant was subject to confiscation of all of his or her property for a specific crime as determined by the court. &  1 = Yes; 0 = No; Non-mention is treated as 0\\
\hline
\url{Confiscated\_property\_amount} & A CONFISCATED\_PROPERTY\_AMOUNT element refers to the amount of confiscated property as determined by the court for a specific crime of a defendant. This element is annotated only when either element PARTIAL\_PROPERTY\_CONFISCATION or element TOTAL\_PROPERTY\_CONFISCATION is assigned a value of 1. & Extracted from the Specific Content of Judicial Documents\\
\hline
\url{Deportation} &  A DEPORTATION element refers to whether the defendant was subject to deportation for a specific crime as determined by the court. &  1 = Yes; 0 = No; Non-mention is treated as 0\\
\hline

\url{Loss\_compensation} & A LOSS\_COMPENSATION element refers to whether the defendant was ordered to make compensation for the economic loss for a specific crime as determined by the court. &  1 = Yes; 0 = No; Non-mention is treated as 0\\
\hline
\url{Compensation\_amount} & A COMPENSATION\_AMOUNT element refers to the amount of making compensation for the economic loss as determined by the court for a specific crime of a defendant. This element is annotated only when element LOSS\_COMPENSATION is assigned a value of 1. & Extracted from the Specific Content of Judicial Documents\\
\hline

\url{Formal\_apology} & A FORMAL\_APOLOGY element refers to whether the defendant was ordered to make a statement of repentance or formal apology for a specific crime as determined by the court. &  1 = Yes; 0 = No; Non-mention is treated as 0\\
\hline

\url{Administrative\_sanction} & An ADMINISTRATIVE\_SANCTION  element refers to whether the defendant was subjected to administrative sanctions by the relevant department for a specific crime as determined by the court. &  1 = Yes; 0 = No; Non-mention is treated as 0\\
\hline

\url{Occupational\_prohibition} & An OCCUPATIONAL\_PROHIBITION  element refers to whether the defendant was subjected to occupational prohibition for a specific crime as determined by the court. &  1 = Yes; 0 = No; Non-mention is treated as 0\\
\hline

\url{Confiscate\_illegal\_income} & A CONFISCATE\_ILLEGAL\_INCOME element refers to whether the defendant was subjected to the confiscation of illegal income for a specific crime as determined by the court. &  1 = Yes; 0 = No; Non-mention is treated as 0\\
\hline
\url{Confiscated\_income\_amount} & A CONFISCATED\_INCOME\_AMOUNT element refers to the amount of confiscated income for a specific crime of the defendant as determined by the court. This element is annotated only when element CONFISCATE\_ILLEGAL\_INCOME is assigned a value of 1. & Extracted from the Specific Content of Judicial Documents\\
\hline
\url{Other\_confiscation} & An OTHER\_CONFISCATION element refers to whether the defendant was subjected to the confiscation of other objects related to a specific crime of the defendant as determined by the court in the aggregated sentencing. &  1 = Yes; 0 = No; Non-mention is treated as 0\\
\hline
\hline
\end{tabularx}

\captionof{table}{\centering List of detailed element information (X).}
\end{minipage}
\end{table}
\clearpage

\begin{table}
\centering
\begin{minipage}{\textwidth}
\newcolumntype{Y}{>{\hsize=.7\hsize\arraybackslash}X}
\newcolumntype{Z}{>{\hsize=1.45\hsize\arraybackslash}X}
\newcolumntype{S}{>{\hsize=0.85\hsize\arraybackslash}X}
\begin{tabularx}{\textwidth}{Y|Z|S}
\hline
\hline
\textbf{Element Name} & \textbf{Explanation} & \textbf{Value Type} \\
\hline
\multicolumn{3}{c}{\textbf{Crime Characteristics (VI)}}\\
\hline

\url{Return\_illegal\_gain} & A RETURN\_ILLEGAL\_GAIN element refers to the order of returning illegal gains and compensations for a specific crime of the defendant as determined by the court. & 1 = Yes; 0 = No; Non-mention is treated as 0\\
\hline
\url{Return\_amount} & A RETURN\_AMOUNT element refers to the amount of returning illegal gains and compensations as ordered by the court for a specific crime of the defendant. This element is annotated only when element RETURN\_ILLEGAL\_GAIN is assigned a value of 1. & Extracted from the Specific Content of Judicial Documents\\
\hline
\url{Other\_nonpenalty\_punishment} & A OTHER\_NONPENALTY\_PUNISHMENT element refers to the order of returning illegal gains and compensations for a specific crime of the defendant as determined by the court. & 1 = Yes; 0 = No; Non-mention is treated as 0\\

\hline
\hline
\end{tabularx}

\captionof{table}{\centering List of detailed element information (XI).}
\end{minipage}
\end{table}
\clearpage

\end{document}